\documentclass{article}
\usepackage{spconf,amsmath,graphicx}
\usepackage{todonotes}
\usepackage{booktabs}
\usepackage{subfigure}
\usepackage{romannum}
\usepackage{graphicx}
\usepackage{soul}
\usepackage{algorithm}
\usepackage[noend]{algpseudocode}
\usepackage{algorithmicx,algorithm}
\usepackage{multirow}
\usepackage[font=small,labelfont=bf]{caption}

\title{Single-shot Channel Pruning Based on Alternating Direction Method of Multipliers}
\name{Chengcheng Li$^{1,*}$, Zi Wang$^{1,*}$, Xiangyang Wang$^2$, Hairong Qi$^1$ 
\thanks{$^*$ With equal contribution. Submitted to the 2019 IEEE International
Conference on Image Processing. Personal use of this material is permitted.
However, permission to reprint/republish this material for advertising or promotional
purposes or for creating new collective works for resale or redistribution
to servers or lists, or to reuse any copyrighted component of this work
in other works must be obtained from the IEEE.}
}
\address{$^1$Department of Electrical Engineering and Computer Science, University of Tennessee, USA \\
$^2$School of Mathematics, Sun Yat-Sen University, China}
\begin{document}
\maketitle

\begin{abstract}
Channel pruning has been identified as an effective approach to constructing efficient network structures. Its typical pipeline requires iterative pruning and fine-tuning. In this work, we propose a novel single-shot channel pruning approach based on alternating direction methods of multipliers (ADMM), which can eliminate the need for complex iterative pruning and fine-tuning procedure and achieve a target compression ratio with only one run of pruning and fine-tuning. To the best of our knowledge, this is the first study of single-shot channel pruning. The proposed method introduces filter-level sparsity during training and can achieve competitive performance with a simple heuristic pruning criterion ($l_1$-norm). Extensive evaluations have been conducted with various widely-used benchmark architectures and image datasets for object classification purpose. The experimental results on classification accuracy show that the proposed method can outperform state-of-the-art network pruning works under various scenarios.
\end{abstract}
\section{Introduction}
\label{sec:intro}
In the past decade, deep convolutional neural networks (DCNN) have achieved significant success in a wide spectrum of applications, such as object classification and detection \cite{he2017mask, schroff2015facenet}, image synthesis \cite{brock2018large, isola2017image, li2018fast}, and reinforcement learning based applications \cite{silver2017mastering,wang2018deep}. 
% With increasing number of parameters and deeper structures \cite{krizhevsky2012imagenet, simonyan2014very, he2016deep}, superhuman performance have been achieved in more and more applications.
However, their property of over-parameterization unavoidably leads to costly computation, memory, and energy consumption, which adds a significant burden on resource-limited devices, such as cars, mobile phones, and wearable devices. Existing studies \cite{sze2017efficient,zhang2018systematic,liu2018efficient, li2016pruning,molchanov2016pruning} have shown that network pruning is an effective method to reduce the model size without much performance degradation.
% (Fig.\ref{fig:intro_prune}). 

% \begin{figure}[htb]
% \centering
% \begin{minipage}{0.9\columnwidth}
% \centering
% \includegraphics[width=\textwidth]{figs/intro_fig.eps}
% \end{minipage}
% \vspace{-0.15cm}
% \caption{Pruning.}
% \label{fig:intro_prune}
% \vspace{-0.4cm}
% \end{figure}
One pioneering work presented weight-based pruning \cite{han2015learning}, which zeros out the weights with the smallest magnitudes and reduces the number of non-zero parameters of AlexNet by a factor of $9\times$ without performance loss after fine-tuning. Several subsequent developments \cite{zhang2018systematic,liu2018efficient,lee2018viterbi} have further pushed the state-of-the-art. Weight-based pruning methods introduce large weight-level sparsity into the neural network, but they do not remove the zero value parameters, so the size of the model remains the same without specialized hardware/software designs.  

Recently, channel-based pruning approaches \cite{li2016pruning,molchanov2016pruning,luo2017thinet,he2017channel} have attracted lots of attention, which remove entire filters as well as the corresponding feature maps, without the requirement of specialized software and hardware. Its typical pipeline contains three steps, 1) pre-training an over-parameterized neural network, 2) pruning least important filters based on a certain criterion, and 3) fine-tuning to alleviate performance degradation. The last two steps are an iterative procedure. 
% The detailed workflow is shown in Fig. \ref{fig:paradigm}(right). 

The very recent work \cite{lee2018snip} presented a single-shot weight pruning method, which can identify important weights of the network before training. After pruning least important weights, the sparse sub-structure is trained in the standard way. This novel pipeline eliminates the need for the complex iterative pruning and fine-tuning. 
% \vspace{-0.2cm}
\begin{figure}[htb]
\centering
\begin{minipage}{0.8\columnwidth}
\centering
\includegraphics[width=\textwidth]{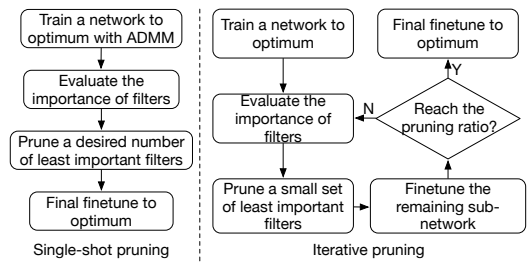}
\end{minipage}
\vspace{-0.15cm}
\caption{Comparison of two paradigms for channel pruning. Left: the proposed single-shot pruning. Right: typical iterative pruning}
\label{fig:paradigm}
% \vspace{-0.3cm}
\end{figure}

Inspired by the single-shot weight-based pruning work in \cite{lee2018snip}, in this paper, we propose a novel single-shot channel pruning approach. In order to improve the performance, we further incorporate alternating direction methods of multipliers (ADMM), which enforces channel-level sparsity \cite{boyd2011distributed, ye2018progressive}. Unlike the typical pruning pipeline which relies on iterative pruning and fine-tuning, the proposed approach needs only one run of pruning and fine-tuning. Given a network architecture, our proposed approach first trains a DCNN with ADMM, then removes the desired number of least important filters with $l_1$-norm criterion, and finally fine-tunes the remaining sub-network. Fig.~\ref{fig:paradigm} presents a comparison between the proposed single-shot channel pruning and typical iterative channel pruning paradigms.            
% proposed approach provides a good start point for the reserved filters to accommodate the pruning structure more smoothly and quickly, so that it can achieve a better performance after pruning and fine-tuning, even without a iterative paradigm. 

Our method is evaluated with two widely-used architectures (LeNet-5 \cite{lecun1998gradient} and AlexNet \cite{krizhevsky2012imagenet}) and two benchmark datasets (MNIST \cite{lecun1998gradient} and Cifar-10 \cite{krizhevsky2009learning}) for object classification purpose. We compare our method with the state-of-the-art channel pruning methods in different scenarios. Experimental results show that the proposed method can significantly outperform state-of-the-art works in the scenario of single-shot pruning. Furthermore, our single-shot pruning method can even outperform the state-of-art methods with iterative pruning.
% our proposed approach achieve state-of-art performance comparing with several important recent channel pruning studies in a pruning and fine-tuning once paradigm. Moreover, we even achieve better performance than the algorithms working on a iterative pruning and fine-tuning paradigm.

\section{Proposed Approach}
\label{sec:approach}
We propose a novel single-shot channel pruning method, built upon \cite{zhang2018systematic}. Given a network architecture, our framework includes three steps, 1) training the DCNN with ADMM which introduces channel-level sparsity, 2) removing the desired number of least important filters, i.e., filters with smallest $l_1$-norm and 3) fine-tuning the remaining sub-network. Compared with existing methods that rely on typical pipeline requiring iterative pruning and fine-tuning, our method only needs one run of pruning and fine-tuning. In the following paragraphs, we present the formulation of channel pruning with ADMM in details. At the end of this section, we summarize our overall pruning procedure in Algorithm \ref{alg:admmc}.

Suppose that we have an N-layer DCNN parameterized by $(W = \{W_i\}_{i=1}^{N})$, where $W_i$ represents the weights and bias of the $i$-th layer. Its associated loss function over the set of training samples $D$ is $C(W, D)$. The pruning problem can then be formulated as (\ref{eq:1}).
\vspace{-0.36cm}
\begin{equation}
\begin{aligned}
& \underset{W}{\text{minimize}}
& & C(W, D) + \sum_{i=1}^N g_i(W_i)\\
& \text{subject to}
& & W_i \in S_i, i=1,...,N,
\label{eq:1}
\end{aligned}
\end{equation}

\noindent where $S_i=\{W_i~|~\text{card}(W_i)\leq l_i\}$ with $l_i$ being the pre-set sparsity level of each layer, and $g_i(\cdot)$ is the indicator function of $S_i$:
\vspace{-0.36cm}
\begin{align*}
g_i(W_i)=\begin{cases}
        0 ~~~~~~~~~~~{\text{if card($W_i$)}} \leq l_i,\\
        +\infty ~~~~~~{\text{otherwise}}.
        \end{cases}
\end{align*}

In weight pruning, the cardinality function card$(W_i)$ returns the number of non-zero elements in $W_i$ \cite{zhang2018systematic}, and weight-level sparsity is introduced by card$(W_i) \leq l_i$. However, in channel pruning, card$(W_i)$ means the number of nonzero filters and $l_i$ is the pre-set filter-level sparsity. 

Taking original LeNet-5 as an example with two convolutional layers. The first convolutional layer has six $5\times 5$ kernels, i.e., $5\times 5\times 6=150$ weight elements, referred to as $W_1$. Assume its pre-set sparsity level is $l_1=2$. For weight-based pruning, $W_1 = \{w_{1,1},\cdots,w_{1,150}\}$, and the indicator function $g_1(W_1)$ equals to 0 if the number of non-zero elements in $W_1$ is less than or equal to 2. However, in channel-based pruning, since we are treating the pruning unit at a higher level, i.e., the filter level, $W_1=\{W_{1,1}, W_{1,2},\cdots,W_{1,6}\}$, with each element of $W_1$ being the filter itself instead of the elements of the filter. So the indicator function would be zero if more than $6-2=4$ filters with all their elements being zero.

% In order to adapt to channel-based pruning, we need to modify this constraint with $||W_i||_0 \leq B_i$, where $||\cdot||_0$ is the $\ell_0$ norm. Note in channel-level pruning, each element in $W_i$ is the parameter corresponding to a filter and $B_i$ is the maximal number of remained filters. Then the indicator function can be rewritten as:
% \begin{align*}
% g_i(W_i)=\begin{cases}
%         0 ~~~~~~~~~~~||W_i||_0 \leq B_i,\\
%         +\infty ~~~~~~{\text{otherwise}}.
%         \end{cases}
% \end{align*}

% QI: Take the example of AlexNet-5 with two convolutional layers, referred to as $W_1$ and $W_3$. Assume the pre-set sparsity level is $l_1=l_3=2$. For weight-based pruning, $W_1 = \{w_{1,0},...,w_{1,149}\}$, the indicator function $g_1(W_1)$ equals to 0 if the number of elements in the $5\times5\times 6$ is less than 2. However, in channel-based pruning, since we are treating the network at a higher level, i.e., the filter level, $W_1=\{W_{10}, W_{11}, ... W_{14}\}$, with each element of $W_1$ being the filter itself instead of the elements of the filter. So the indicator function would be zero if more than 2 filters with all their elements being zero.

% Qi: Also, redefine $l_i$, in weight-level pruning, it means ...; in channel-level pruning, it means the maximal number of nonzero filters. The indicator function would be of the same format, just that the definition of $l_i$ and $card(W_i)$ are different. So we don't need to show $g_i(W_i)$ again.

Since the second term of (\ref{eq:1}) is not differentiable, ADMM \cite{boyd2011distributed} can be adopted to solve this problem. Apparently, (\ref{eq:1}) is interchangeable with its ADMM form (\ref{eq:2}).
\vspace{-0.2cm}
\begin{equation}
\begin{aligned}
& \underset{W}{\text{minimize}}
& & C(W,D) + \sum_{i=1}^N g_i(Z_i) \\
& \text{subject to}
& & W_i = Z_i, i=1,...,N.
\label{eq:2}
\end{aligned}
\end{equation}
The augmented Lagrangian of  (\ref{eq:2}) is:
\vspace{-0.2cm}
\begin{multline*}
L_{\rho}(W,Z,\Lambda) = C(W,D) + \sum_{i=1}^N g_i(Z_i) \\
+\sum_{i=1}^N tr\left[ \Lambda_i^T(W_i-Z_i) \right]+ \sum_{i=1}^N\frac{\rho_i}{2}||W_i-Z_i||_F^2,
\end{multline*}
\noindent where $\Lambda_i$ is the dual variable, $tr(\cdot)$ is the trace, $\rho=\{\rho_i,...,\rho_N\}$ are positive penalty parameters, and $||\cdot||_F$ is the Frobenius norm. By using the scaled dual variable $U_i=(1/\rho_i)\Lambda_i$, the augmented Lagrangian can be rewritten as: 
\vspace{-0.2cm}
\begin{multline*}
L_{\rho}(W,Z,U) = C(W,D) + \sum_{i=1}^N g_i(Z_i) \\
+\sum_{i=1}^N\frac{\rho_i}{2}||W_i-Z_i+U_i||_F^2.
\end{multline*}
According to the ADMM method, the above problem can be divided into two subproblems (\ref{eq:3}) and (\ref{eq:4}):
\vspace{-0.2cm}
\begin{equation}
\begin{aligned}
& \underset{W}{\text{minimize}}
& & C(W,D) + \sum_{i=1}^N \frac{\rho_i}{2}||W_i-Z_i^k+U_i^k||_F^2.
\label{eq:3}
\end{aligned}
\end{equation}
\vspace{-0.2cm}
\begin{equation}
\begin{aligned}
& \underset{W}{\text{minimize}}
& \sum_{i=1}^N g_i(Z_i) + \sum_{i=1}^N \frac{\rho_i}{2}||W_i^{k+1}-Z_i+U_i^k||_F^2.
\label{eq:4}
\end{aligned}
\end{equation}
The first term of (\ref{eq:3}) is the loss function and the second term can be considered as a special regularizer that is differentiable. According to \cite{boyd2011distributed,zhang2018systematic}, the globally optimal solution of (\ref{eq:4}) can be explicitly derived as:
\vspace{-0.1cm}
\begin{equation*}
\begin{aligned}
Z_i^{k+1} = \Pi_{S_i}(W_i^{k+1} + U_i^k),
\end{aligned}
\end{equation*}
\noindent where $\Pi_{S_i}(\cdot)$ denotes the Euclidean projection onto $S_i$. $Z_i^{k+1}$, where $k$ is the index of iterations, can be obtained by preserving the $B_i$ filters with the largest $l_1$ norm and zeroing out the rest in the $i$th layer. After solving (\ref{eq:3}) and (\ref{eq:4}), we update $U_i^{k+1}$ as $U_i^k+W_i^{k+1}-Z_i^{k+1}$. Overall, the whole problem can be solved by iteratively updating $W$, $Z$, and $U$. 

We summarize our proposed approach in Algorithm \ref{alg:admmc}.

\begin{algorithm}
\caption{The proposed ADMM-based channel pruning approach for classification tasks}
\label{alg:admmc}
{\bf Input:} 
Initialized $W_i$, learning rate $\alpha$, tolerance thresholds $\epsilon_i$, pruning rates $p_i\%$, ADMM scaling factor $\rho$, ADMM update interval $M$.\\
{\bf Output:} 
A pruned DCNN.
\begin{algorithmic}[1]
\State Train the DCNN with cross-entropy loss and $l_{2}$ regularization.
\State Project $Z_i^1$ onto $S_i$ by zeroing out the $p_i\%$ filters with the smallest $l_1$ norm in the $i$th layer.
\State Initialize $U_i^1 = 0$, which has the same dimension as $Z_i$. 
\State k=0.
\While{$||W_i^{k+1}-Z_i^{k+1}||_F^2>\epsilon_i$,$||Z_i^{k+1}-Z_i^k||_F^2>\epsilon_i$}
  \State k=k+1.
  \While{iter $\leq M$}
    \State Update $W_i$ with backpropagation gradient ($\Delta W_i$) and ADMM regularization: $W_i^{k} = \alpha(W_i^k - \Delta W_i^k + \rho(W_i^k - Z_i^k + U_i^k))$.
  \EndWhile
  \State $W_i^{k+1}=W_i^k$.
  \State Update $Z_i^{k+1} = W_i^{k+1} + U_i^k$.
  \State Project $Z_i^{k+1}$ onto $S_i$.
  \State Update $U_i^{k+1} = U_i^k + W_i^{k+1} - Z_i^{k+1}$
\EndWhile
\State Prune the $p_i\%$ filters with the smallest $l_1$ norm and finetune the DCNN to optimum.
\end{algorithmic}
\vspace{-0.1cm}
\end{algorithm}
\vspace{-0.1cm}
\section{Experiments and Results}
\label{sec:exp}
We conduct extensive experiments to evaluate the proposed single-shot channel pruning approach. In the following, after a description of the experimental setup, we compare the performance of the proposed method with several state-of-the-art works in different scenarios. We conclude this section by in-depth investigation on the mechanisms of ADMM. 
\vspace{-0.1cm}
\subsection{Experimental Setup}
\label{sec:setup}
We evaluate our proposed algorithm on two popular network structures, LeNet-5 and AlexNet, using two benchmark datasets, MNIST and Cifar-10, for object classification purpose. The LeNet-5 network has two convolutional layers, with 20 and 50 filters, respectively. Note that we increase the number of filters in each convolutional layer from the original design in order to better show the effect of pruning. The AlexNet has $5$ convolutional layers, with 64, 192, 394, 256, 256 filters, respectively. 

In each experiment, we train two sets of neural networks from scratch with the identical sets of hyperparameters (SGD optimizer, a learning rate of $0.0001$, and $l_2$ regularization). The difference is, one is with ADMM and the other is without ADMM for pruning methods that do not need ADMM.
Without ADMM, the accuracies of the pre-trained networks are $99.1\%$ and $77.6\%$ for LeNet-5 on MNIST, and AlexNet on Cifar-10, respectively.  
% When applying channel-level ADMM, as the sparsity ratio gets higher, the models' performance would be slightly reduced due to the extra constraints introduced by ADMM.
% but this deficiency can be mitigated after pruning and fine-tuning.
After training, we prune a certain percent of filters and fine-tune the remaining sub-structures for 100 epochs, which can guarantee convergence for all cases. 
%The ADMM procedure can be implemented simply with adding a regularizer and our preliminary study shows that the best performance can be achieved with a scalar of $0.001$. 

We compare our method with state-of-the-art channel pruning works, including, 
1. {\bf minimum weight} \cite{li2016pruning}: ranking filters with $l_1$-norms of the kernel weights. 2. {\bf mean activation} \cite{polyak2015channel}: ranking filters with the mean values of the $l_1$-norms of the activation maps. 3. {\bf Taylor expansion} \cite{molchanov2016pruning}: ranking filters according to saliency-based criterion, Taylor expansion. 4. {\bf ADMM-weight} \cite{zhang2018systematic}: the weight-based ADMM approach is used in the training process and filters are removed in the pruning phase. 5. {\bf random}: pruning randomly-selected filters. Most of these works follow the typical pruning pipeline but with different ranking criteria on the filters. 

\vspace{-0.1cm}
\subsection{Comparison with Single-Shot Pruning}
\label{singleshotcompare}
We compare the performance of our approach with various state-of-the-art works listed in Section \ref{sec:setup}, in the scenario of single-shot pruning by removing a desired number of filters and fine-tuning in one run. 
Since classification on MNIST is not a challenging task, our preliminary study shows that there is no distinguishable performance difference between different methods when the pruning ratio is less than $50\%$. %When pruning ratio is too big, performance degradation is significantly beyond the acceptable tolerance. 
Hence, we only report the performance comparison with the pruning ratio in the range of $[50\%, 95\%]$. Similarly, for AlexNet on Cifar-10, the reported pruning ratio is in the range of $[12.5\%, 87.5\%]$.

% \begin{table}[htb]
% \centering
% \begin{tabular}{ccc}
% \hline
%  & Pruning rate & Reduction ratio \\
% \hline
% \hline
% \multirow{ 4}{*}{LeNet-5} & 50\% &   2x \\
% & 75\% &   4x \\
% & 87.5\% &   8x \\
% & 95\% &   20x \\
% \hline
% \hline
% \multirow{ 5}{*}{AlexNet} & 12.5\% &   1.14x \\
% & 25\% &   1.33x \\
% & 50\% &   2x \\
% & 75\% &   4x \\
% & 87.5\% &   8x \\
% \hline
% \end{tabular}
% % \vspace{-0.15cm}
% \caption{Pruning configurations.}
% \label{tab:prune_config}
% % \vspace{-0.5cm}
% \end{table}

\begin{figure}[htb]
\centering
\begin{minipage}{\columnwidth}
\centering
\subfigure[LeNet-5 on MNIST]{
\includegraphics[width=0.485\textwidth]{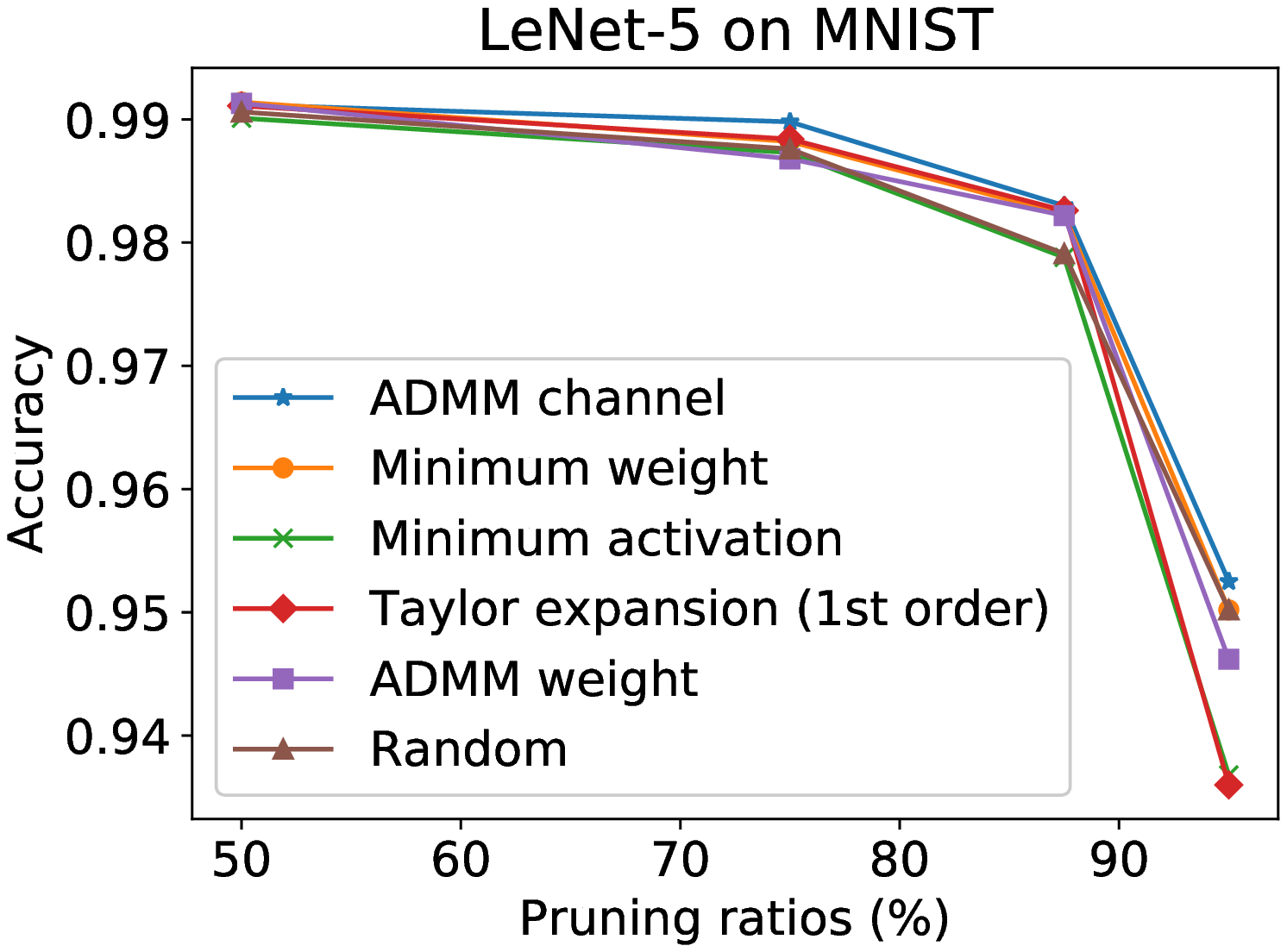}}
\subfigure[AlexNet on Cifar-10]{
\includegraphics[width=0.485\textwidth]{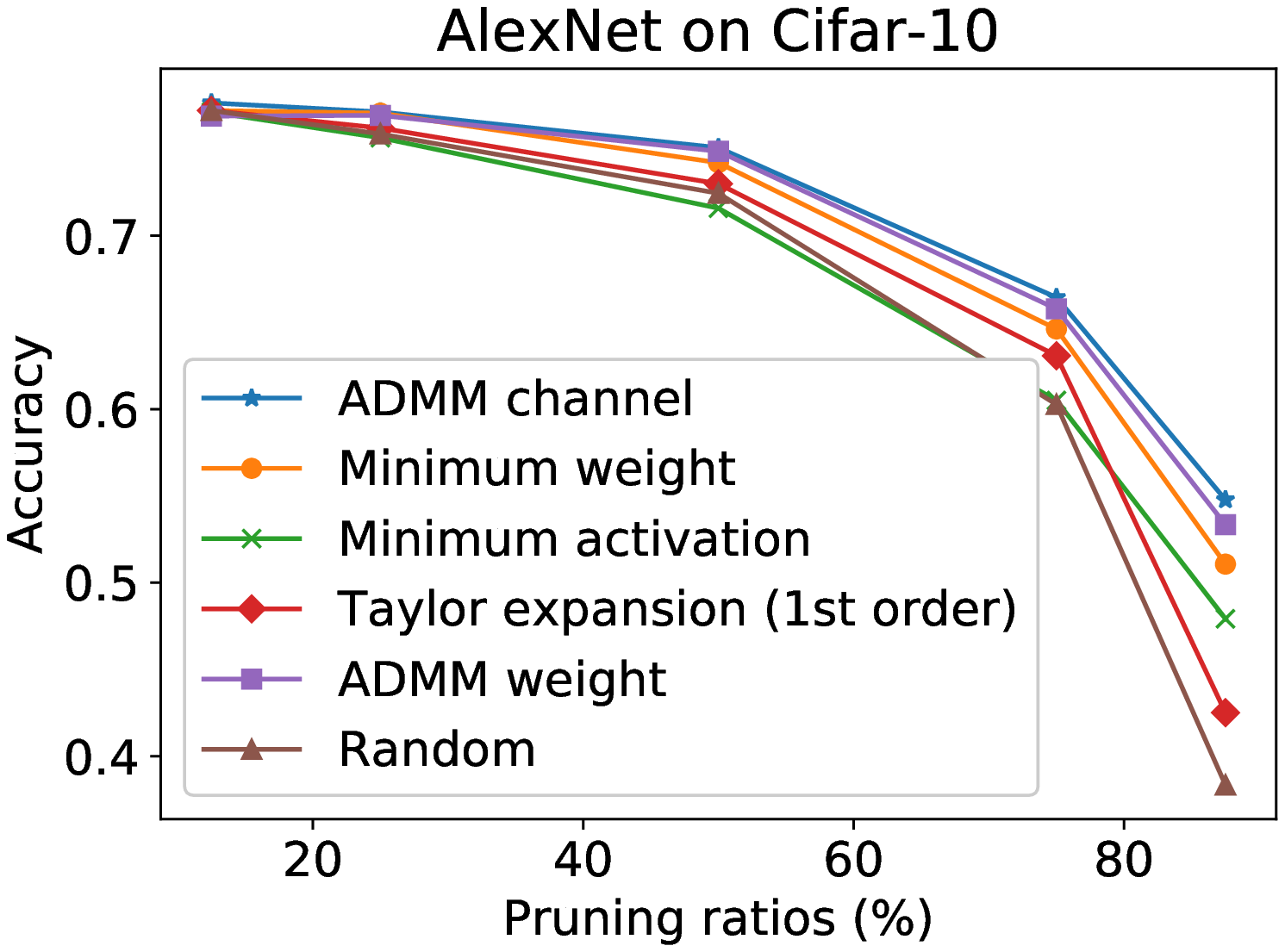}}
\end{minipage}
\vspace{-0.35cm}
\caption{Performance comparison with single-shot pruning.}
\label{fig:performance}
% \vspace{-0.4cm}
\end{figure}

The results are shown in  Fig.~\ref{fig:performance}, where we observe that the proposed method can achieve the best performance with all pruning ratios except for $50\%$, where the minimum weight achieves the best accuracy (99.14\% vs. ours 99.12\%). In specific, with $50\%$ pruning ratio, there is no noticeable performance degradation and all methods achieve comparable performance. One possible interpretation is LeNet-5 is a significantly overparameterized structure for MNIST classification task. A smaller structure with $50\%$ less filters can achieve comparable performance.
% \vspace{-0.35cm}

It is clear that, with AlexNet on Cifar-10, our approach achieves the best performance on all pruning ratios. As the pruning ratio increases, the proposed approach shows a larger performance margin over other approaches. %Besides, the superiority of our method is more obvious as more filters are pruned. These results indicate existing pruning methods probably rely on the iterative pruning/ finetuning procedure. 

\vspace{-0.1cm}
\subsection{Comparison with Iterative Pruning}
In this set of experiments, we compare the proposed single-shot pruning with state-of-the-art iterative pruning approaches. Since Taylor expansion (TE) has achieved the best performance \cite{molchanov2016pruning} in existing iterative pruning approaches, we only use TE as the criterion for iterative pruning. Since the results in Fig.~\ref{fig:performance} showed that LeNet-5 on MNIST is not as challenging as AlexNet on Cifar-10, in the subsequent experiments, we only report results with AlexNet on Cifar-10.

In Section \ref{singleshotcompare}, we prune the same percent of filters from each layer. In this challenging scenario, we allow the pruning ratio of different layers to be different. The actual pruning ratio used at each layer is determined through an empirical study. Then we input this information in the pre-training phase for the purpose of ADMM and directly prune the target ratio of filter for each layer. Finally, we fine-tune both networks with $100$ epochs, which is enough for them to get converged. For the TE method, in each run, we prune $10$ filters and fine-tune the network with $500$ updates with a batch size of $50$. We consider two cases for the TE method after the iterative pruning and fine-tuning, as suggested in \cite{molchanov2016pruning}, 1) typical pipeline without extra fine-tuning and 2) extra fine-tuning of $100$ epochs after the typical pruning/fine-tuning procedure. 

%Note that our proposed method does not need iterative pruning and fine-tuning, but we allow Taylor expansion method does iterative pruning and fine-tuning. 

Experimental results with different pruning ratios are illustrated in Table \ref{tab:admm_te}. %We can see extra fine-tuning (FT) can improve the performance of Taylor expansion significantly. 
It is clear that our proposed method outperforms the state-of-the-art iterative channel pruning method for all pruning ratios. It is worth emphasizing that, in this set of experiments, we are comparing the proposed method using just single iteration of pruning and fine-tuning with TE method that employ iterative pruning and fine-tuning.%It is worth noting that compared with Taylor expansion method, the proposed method can save lots of computation in fine-tuning.
% \vspace{-0.1cm}
\begin{table}[htb]
\centering
\begin{tabular}{c|ccc}
\hline
Ratio & ADMM & TE (No extra FT) & TE (Extra FT) \\
\hline
50\% & {\bf 77.17\%} & 73.06\% & 75.47\% \\
\hline
75\% & {\bf 72.04\%} & 62.72\% & 70.03\% \\
\hline
87.5\% & {\bf 64.17\%} & 51.83\% & 60.92\% \\
\hline
\end{tabular}
\vspace{-0.2cm}
\caption{Performance comparison with iterative pruning approach.}
\label{tab:admm_te}
\end{table}
% \vspace{-0.4cm}
% prune 87.5\%: ADMM: 64.17\%, nvidia: 51.83\% (before), 60.92\% (after 50 epo finetune lr=1e-4)
\subsection{Visualizing Characteristics of ADMM}
\begin{figure}[htb]
\centering
\begin{minipage}{\columnwidth}
\centering
\subfigure[LeNet-5 on MNIST]{
\includegraphics[width=0.485\textwidth]{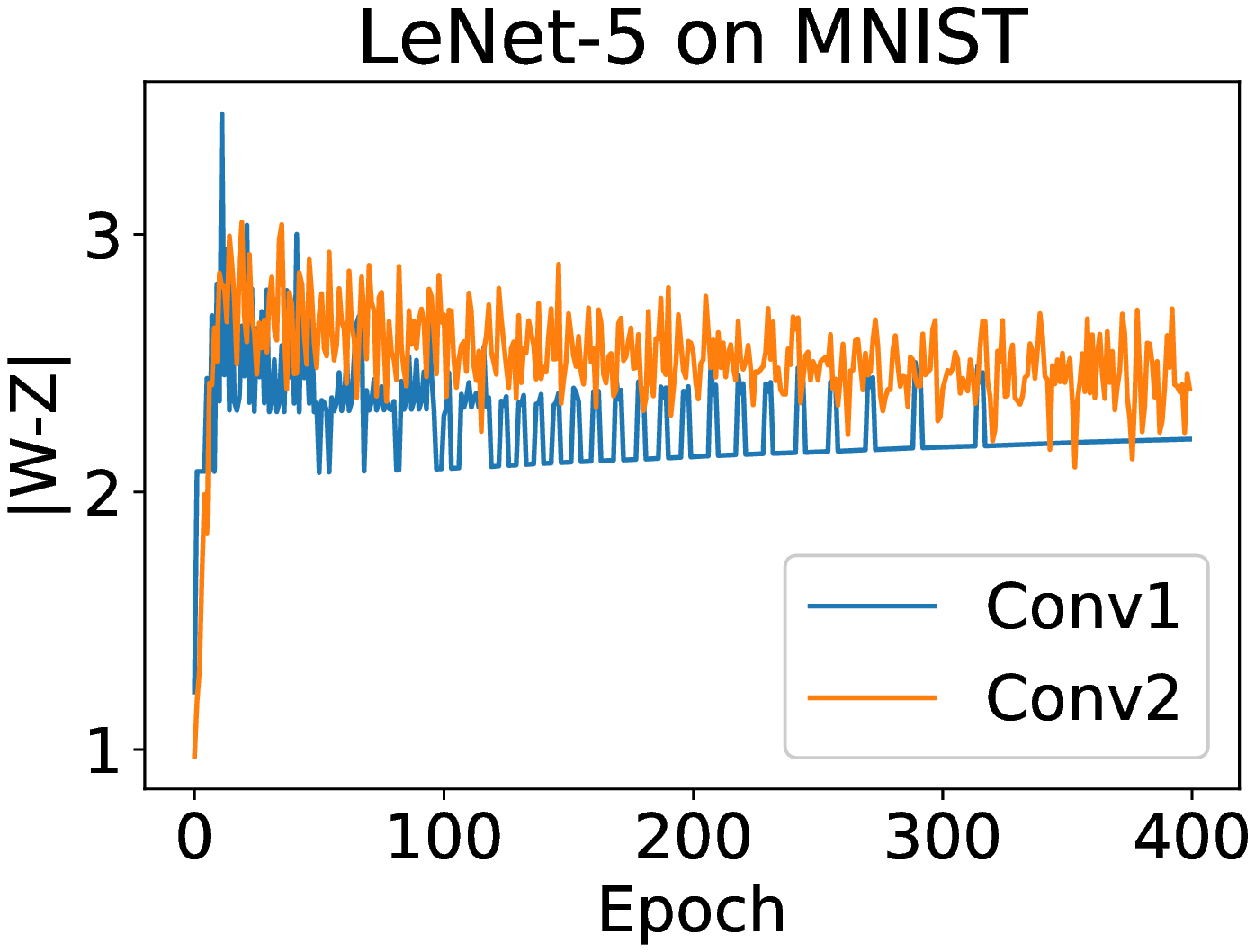}}
\subfigure[AlexNet on Cifar-10]{
\includegraphics[width=0.485\textwidth]{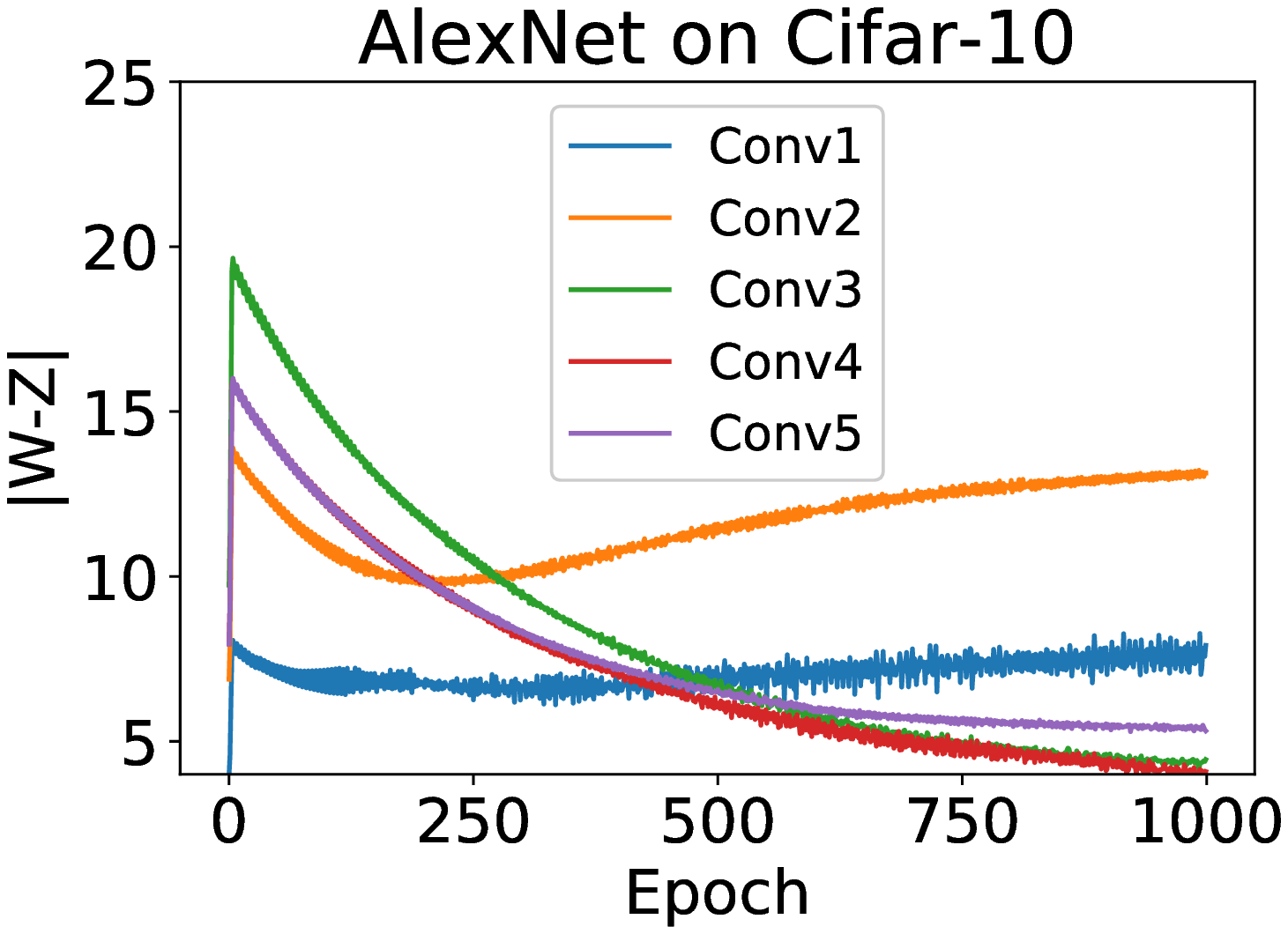}}
\end{minipage}
\vspace{-0.3cm}
\caption{Distance between W and Z during training.}
\label{fig:norm}
% \vspace{-0.5cm}
\end{figure}

In this section, we conduct in-depth study on the effect of ADMM on  network performance. We first study the evolution pattern of sparsity of filters by analyzing the Euclidean distance between filter weights $W$ and the corresponding sparsified mask $Z$ for each layer during training. Results are shown in Fig.~\ref{fig:norm}. It is observed that in the shallower layers, the distance remains approximately the same (Conv 1 of LeNet-5 and AlexNet), or increases slightly (Conv 2 of AlexNet). This observation is consistent with the hypothesis that weights in the shallower layers play more important roles for feature extraction, thus not easily sparsified. However, in the deeper layers (Conv 2 of LeNet-5, Conv 3, 4 and 5 of AlexNet), the distance $|W-Z|$ decreases significantly during training.
% \vspace{-0.2cm}

\begin{figure}[htb]
\centering
\begin{minipage}{\columnwidth}
\centering
\subfigure[LeNet-5 on MNIST]{
\includegraphics[width=0.485\textwidth]{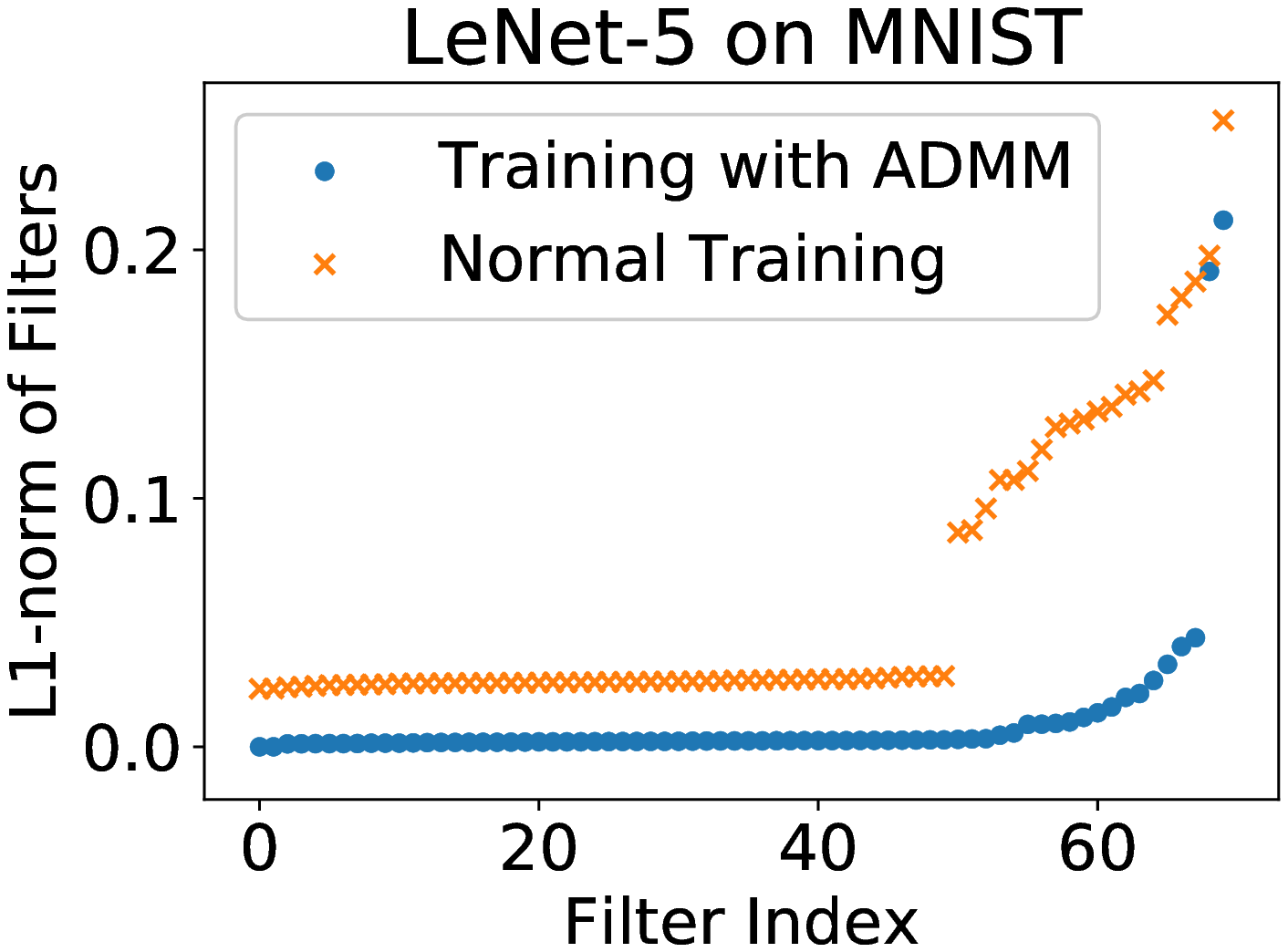}}
\subfigure[AlexNet on Cifar-10]{
\includegraphics[width=0.485\textwidth]{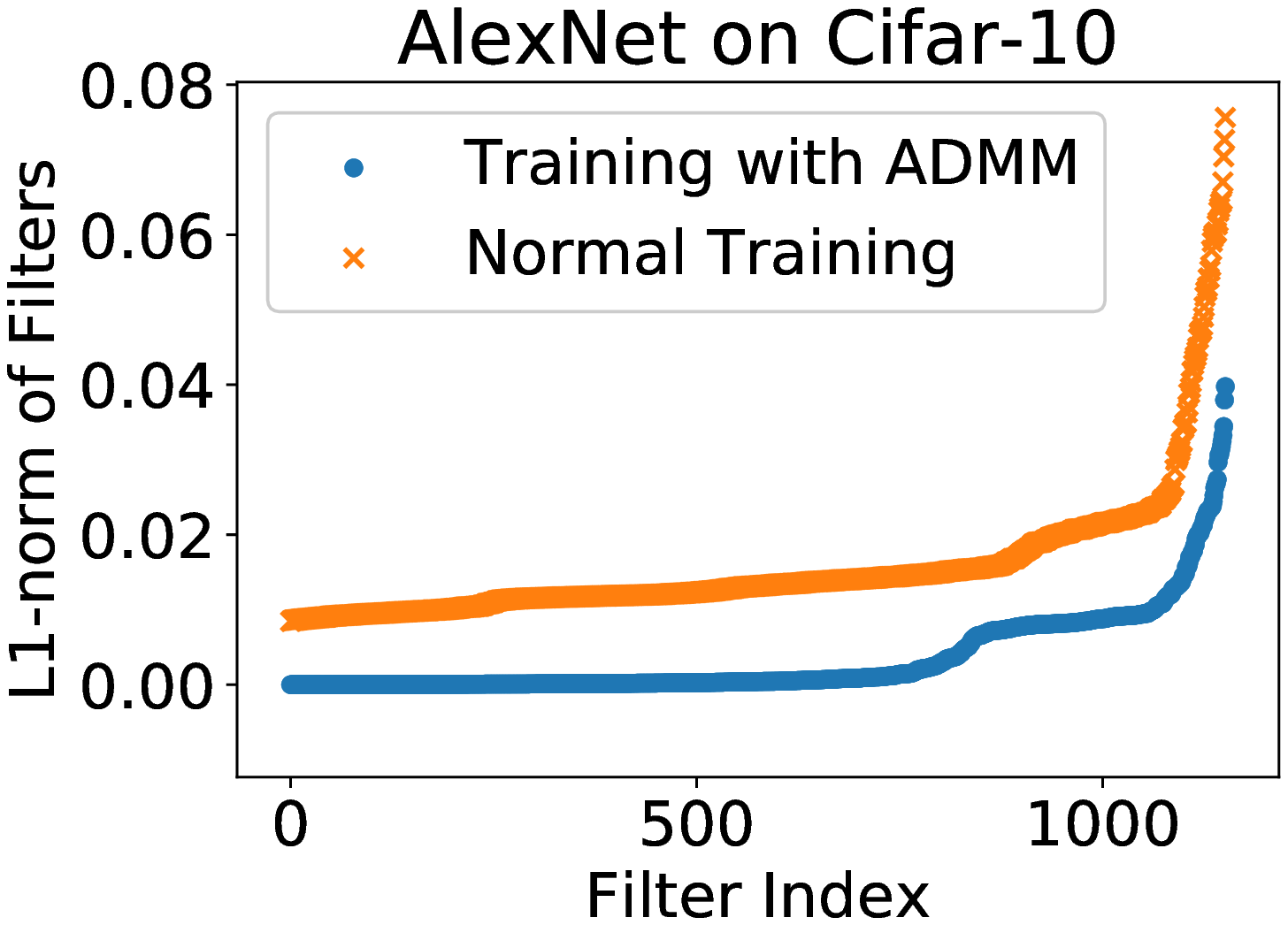}}
\end{minipage}
\vspace{-0.4cm}
\caption{Illustration of $l_1$-norm of all filters.}
\label{fig:norm_dist}
% \vspace{-0.4cm}
\end{figure}

We further visualize the $l_1$ norm of each filter in the pretrained networks to analyze the distribution of the magnitude of filters, as shown in Fig.~\ref{fig:norm_dist}. It is clear that with ADMM, the $l_1$ norms of more filters become very close to zero as compared with normal training. These results verify that ADMM can introduce considerable channel-level sparsity to DCNN.
% \vspace{-0.4cm}

\begin{figure}[htb]
\centering
\begin{minipage}{\columnwidth}
\centering
\subfigure[LeNet-5 on MNIST]{
\includegraphics[width=0.485\textwidth]{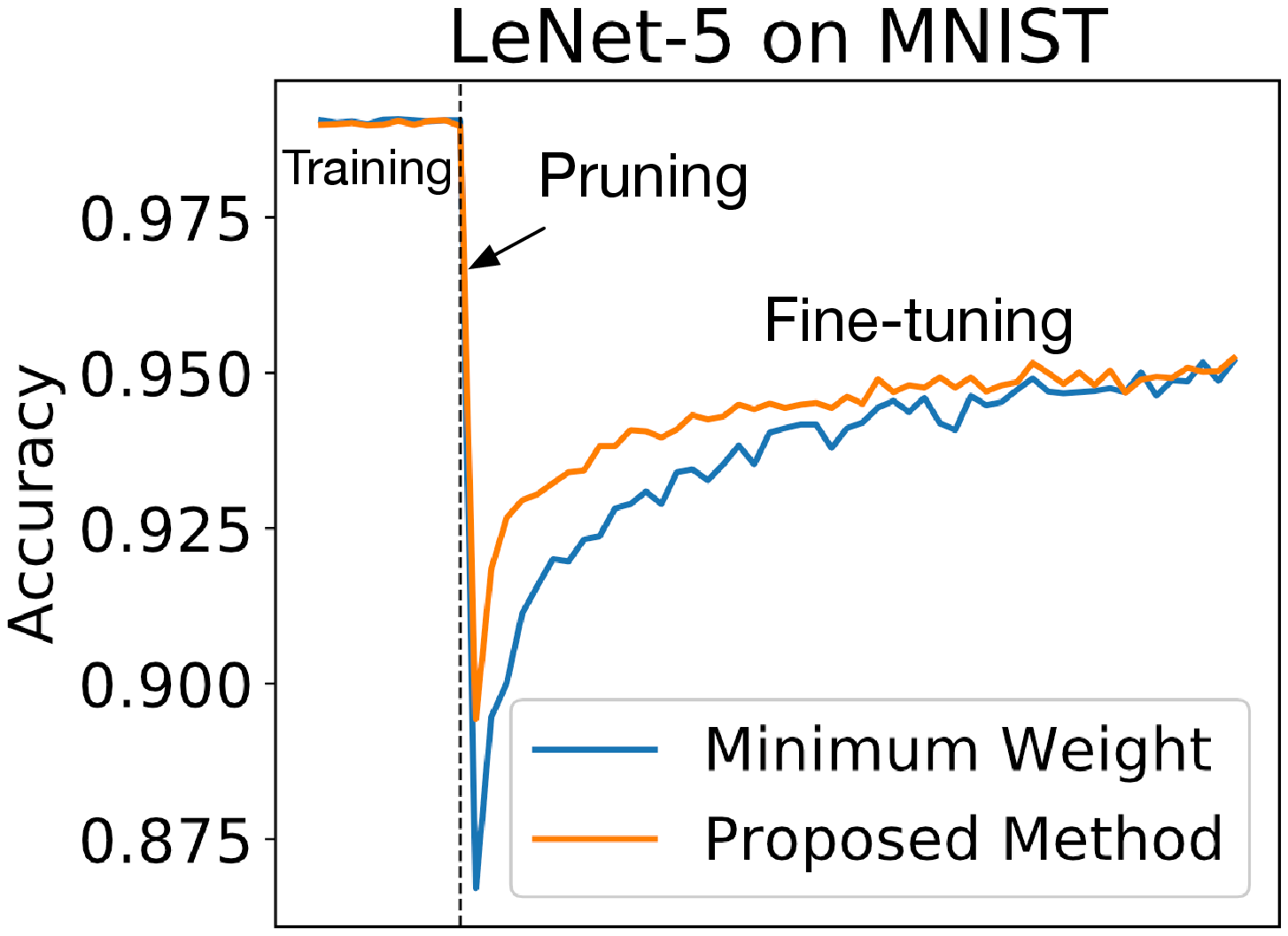}}
\subfigure[AlexNet on Cifar-10]{
\includegraphics[width=0.485\textwidth]{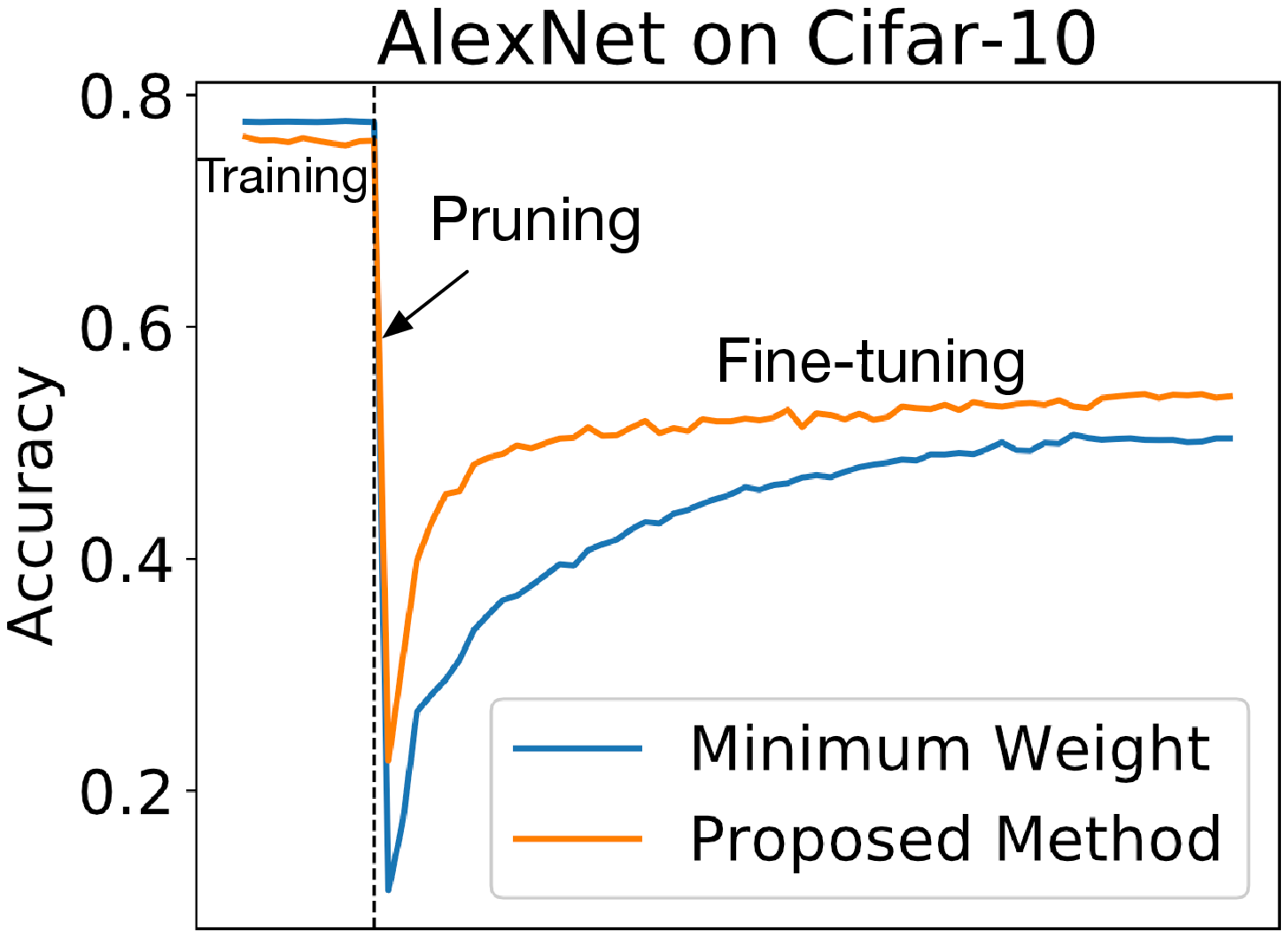}}
\end{minipage}
\vspace{-0.4cm}
\caption{Performance curve during training, pruning and fine-tuning with single-shot scenario.}
\label{fig:pr_performance}
% \vspace{-0.4cm}
\end{figure}

Finally, we illustrate the evolution pattern of classification accuracy during the three stages of pretraining, pruning, and fine-tuning, as shown in Fig.~\ref{fig:pr_performance}. Even through the pretrained networks trained with ADMM achieve a slightly lower accuracy than those without ADMM, after the pruning desired number of filters, the networks pretrained with ADMM result in higher accuracy than those without ADMM. This trend is constantly preserved even in the fine-tuning stage. 
% \vspace{-0.4cm}

% In summary, our proposed ADMM-based channel pruning approach can significantly introduce filter-level sparsity, which can provide a better starting point for the reserved filters after pruning. 
% \vspace{-0.3cm}
\section{Conclusions}
\label{sec:conclusion}
In this paper, we proposed a novel single-shot channel pruning approach that introduces ADMM in training to achieve channel-level sparsity. During pruning, a desired pruning ratio of filters with the smallest $l_1$ norms are removed, and fine-tuning is applied to compensate for performance loss. The proposed method has been evaluated extensively with various widely-used network structures and datasets. Experimental results showed that our method outperforms state-of-the-art works in both single-shot and iterative pruning scenarios.
% Thorough analysis has been conducted to achieve a deep understanding of the proposed method.
\bibliographystyle{IEEEbib}
\bibliography{refs}
\end{document}